\documentclass[11pt]{article}

\usepackage[utf8]{inputenc}
\usepackage[T1]{fontenc}
\usepackage[margin=1in]{geometry}
\usepackage{graphicx}
\usepackage{booktabs}
\usepackage{amsmath}
\usepackage{placeins}
\usepackage{amssymb}
\usepackage{float}
\usepackage[colorlinks=true,linkcolor=blue,citecolor=blue,urlcolor=blue]{hyperref}
\usepackage[round]{natbib}
\usepackage{xcolor}
\usepackage[normalem]{ulem}
\usepackage{authblk}
\newif\ifshowchanges
\showchangesfalse
\ifshowchanges
  \newcommand{\added}[1]{\textcolor{blue}{#1}}
  \DeclareRobustCommand{\deleted}[1]{\textcolor{red}{\sout{#1}}}
\else
  \newcommand{\added}[1]{#1}
  \DeclareRobustCommand{\deleted}[1]{}
\fi

\title{Beyond Persona Prompting: Latent-Conditioned Language Models as Psychometrically Valid Synthetic Respondents\deleted{.}}

\author[1,*]{Nicola Milano}
\author[1]{Davide Marocco}
\affil[1]{University of Naples Federico II, Natural and Artificial Cognition Laboratory ``Orazio Miglino'', Department of Humanities}
\affil[*]{Corresponding author: Nicola Milano, \href{mailto:nicola.milano@unina.it}{nicola.milano@unina.it}}
\date{}

\begin{document}

\maketitle

\begin{abstract}
Large language models have been proposed as "silicon samples" --- synthetic respondents that stand in for human participants in survey and psychometric research. Existing approaches mostly prompt a model with a persona description and read back answers, which we show reproduces human factor structure only weakly. Instead, we keep an instruction-tuned model frozen and train a soft prefix and low-rank adapters that inject a continuous five-dimensional Big Five latent score obtained from real human data. We read each Likert response directly from the model's next-token digit logits, and we build a synthetic population by sampling latents from the empirical covariance of human factor scores. The resulting silicon population attains fair factor congruence with the human five-factor structure on held-out items (Tucker congruence $\approx$ 0.85), transfers to a different inventory with different items and a different item format (IPIP$\rightarrow$BFI-2 $\approx$ 0.80 Tucker congruence\added{, slightly below the conventional .85 threshold for fair similarity}), and preserves \deleted{the inter-factor correlation matrix (scale-score correlations r $\approx$ 0.95 with humans)}\added{the sign pattern of the inter-factor correlations ($r \approx 0.95$ between silicon and human inter-factor correlations)}. Most importantly, the population reproduces emergent, non-injected nomological links: administered four never-seen, non-personality instruments (grit, self-esteem, leadership emergence, procrastination), it recovers the literature-documented \deleted{Big Five correlates}\added{dominant Big Five correlate} with the correct sign in every domain \deleted{and the correct dominant trait under multivariate regression in the published pattern}\added{and, under multivariate regression, isolates the theoretically dominant trait in every checkpoint for grit and leadership while reproducing the weaker multi-trait patterns reported for self-esteem and procrastination} --- even though those domain$\leftrightarrow$trait relationships appear nowhere in the injected covariance. All results are obtained with two open-source models, Qwen2.5-1.5B-Instruct and Llama-3-8B-Instruct, running locally on a single consumer-grade GPU with a relatively small number of trainable parameters.
\end{abstract}

\section{Introduction}

A \textbf{silicon sample} uses a language model to generate synthetic respondents that stand in for a target human population \citep{argyle2023}. The appeal for behavioral research is concrete: synthetic respondents promise to lower the cost and time of data collection, to pilot instruments before fielding them on humans, to reach populations that are small, sensitive, or hard to recruit, and — most ambitiously — to serve as a testbed for \textit{scale development}, where a new questionnaire could be trialed on a synthetic pool before any human sees it \citep{grossmann2023,dillion2023}. Each of these uses treats the synthetic sample as a measurement instrument, and that framing raises a measurement question that surface plausibility does not answer: do synthetic respondents reproduce the \textit{psychometric structure} of the construct being measured, or only its marginal statistics? A model can output entirely plausible Likert numbers whose item means match a human sample, yet whose covariance does not factor into the same latent dimensions. For every use that motivates silicon sampling in the first place — factor analysis, scale authoring, criterion prediction — it is the covariance structure, not the item means, that carries the information.

This distinction is where existing evidence stops short. The silicon-sampling literature has largely validated synthetic respondents at the level of \textit{aggregate response distributions}: whether a demographically conditioned model reproduces the marginal opinion breakdown of a subpopulation \citep{argyle2023,santurkar2023}, or whether simulated agents reproduce the average outcome of a classic experiment \citep{aher2023}. In parallel, a line of work in ``LLM personality'' shows that prompting can shift a model's \textit{mean trait scores} in the intended direction and that these scores can be read off standard inventories \citep{serapiogarcia2023,jiang2023,miotto2022,pellert2024}. What neither line establishes is that a synthetic \textit{population} reproduces the inter-item covariance, factor structure, and nomological network that operationally define a psychological construct \citep{cronbachmeehl1955}. There is moreover direct reason for caution: self-report personality measurements taken from LLMs are sensitive to prompt phrasing and option ordering \citep{gupta2024}, and models exhibit a human-like social-desirability bias that inflates desirable traits once they detect they are being profiled \citep{salecha2024}. Mean-level trait recovery, in short, is both a weaker and a less trustworthy criterion than the structural one that scale developers actually depend on.

We study the structural question for the \textbf{Big Five} personality dimensions (Openness, Conscientiousness, Extraversion, Agreeableness, Neuroticism). Rather than describe a persona in the prompt, we condition a frozen instruction model on a \textit{continuous five-dimensional trait latent} through a small soft prefix and low-rank adapters, read each Likert answer directly off the model's own next-token digit logits (which preserves its language ability), and build a synthetic population by \textbf{sampling} — not taking the argmax of — those response distributions for latents drawn from the empirical human factor-score covariance. Sampling supplies the item-unique variance that psychometric structure lives in, and drawing latents from the human covariance lets the synthetic population reproduce the inter-factor correlations. We then evaluate the population not by whether its trait scores point the right way, but by whether it reproduces human factor structure on \textit{held-out items and a second inventory}, and whether it recovers documented Big Five correlates of \textit{non-personality} instruments it was never trained on.

Our contribution is this method together with a battery of validity evidence:

\begin{enumerate}
\item \textbf{Method.} A \textit{latent-conditioned} respondent simulator: a frozen instruction LM whose Likert answers are read from the model's own digit-token logits (preserving language ability), conditioned on a continuous 5-D Big Five latent through a small soft prefix and low-rank adapters, with a synthetic population built by \textbf{sampling} responses for latents drawn from the empirical human factor-score covariance.
\item \textbf{Structural validity.} The synthetic population attains fair factor congruence with the human five-factor structure on items the model never trained on (mean Tucker congruence $\approx$ 0.85, Neuroticism the weakest factor), and this transfers across inventories and item formats (IPIP$\rightarrow$BFI-2 $\approx$ 0.80), \deleted{preserving the inter-factor correlation matrix}\added{preserving the sign pattern of the inter-factor correlations}.
\item \textbf{Criterion/nomological validity.} The population reproduces documented Big Five correlates of \textit{non-personality} instruments (grit, self-esteem, leadership emergence, procrastination) that are absent from the injected covariance. Beyond the correlation sign pattern, a discriminant multiple regression isolates the theoretically dominant trait — cleanly for grit and leadership — evidence that the simulator has learned construct$\rightarrow$response behavior rather than replaying an injected correlation matrix or an evaluative halo.
\item \textbf{Robustness.} Replication across five seeds\added{ per backbone} and a second, larger backbone (Llama-3-8B), together with held-out temperature calibration and silicon generative (sampling) stability, establishes that the headline numbers are not single-run or hyperparameter artifacts.

\end{enumerate}

We position this against prompt-persona approaches \citep{serapiogarcia2023,jiang2023}, which we include as zero-shot baselines and over which trained latent conditioning roughly doubles factor congruence on held-out items, and against the broader debate on whether LLMs can substitute for human participants \citep{dillion2023,bisbee2024}. We are also explicit about the limit of the design: the conditioning latents are CFA factor scores derived from the training inventory and the inter-factor covariance is injected by construction, so a degree of circularity is built in — which is precisely why the emergent, non-injected criterion validity of \S\ref{sec:criterion} carries the weight of the argument.

\section{Related work}

\subsection{Silicon samples: language models as synthetic respondents}

The proposal to use language models as stand-ins for human respondents originates with \citet{argyle2023}, who condition GPT-3 on demographic backstories and show it reproduces aggregate response patterns of the corresponding subpopulations — a property they term \textit{algorithmic fidelity}. The idea has since been pursued across disciplines: as \textit{homo silicus} in economics, where models given endowments and preferences reproduce classic behavioral-economics findings \citep{horton2023}; as ``Turing experiments'' that replicate the average outcomes of canonical psychology and psycholinguistics studies \citep{aher2023}; as memory-augmented generative agents that enact believable day-to-day behavior \citep{park2023}; and as a general prospect for social-science research \citep{grossmann2023}. A parallel strand probes whose opinions a model actually reflects, finding systematic misalignment between model outputs and the demographic groups they are meant to represent \citep{santurkar2023}. Critical work cautions that synthetic samples can distort distributions and correlations relative to humans and warns against treating them as drop-in replacements \citep{bisbee2024,dillion2023}. What unites these efforts — supportive and critical alike — is that they evaluate fidelity at the level of \textit{aggregate or marginal} response patterns: the marginal breakdown of an opinion item, or the mean outcome of an experiment. The quantity a scale developer cares about — whether the synthetic responses factor into the same latent structure as human data — is left untested. We take that structural reproduction, on held-out items and across inventories, as the object of study.

\subsection{Simulating personality and personas in language models}

A second line of work asks whether LLMs can be made to express a target \textit{personality}. Prompting a model with a persona or a trait description reliably shifts its self-reported Big Five scores in the intended direction, and these scores can be elicited with standard human inventories \citep{serapiogarcia2023,jiang2023,miotto2022}; broader psychometric batteries extend the same idea to values, clinical, and interpersonal scales \citep{pellert2024,huang2024}. These studies establish that trait \textit{scores} are steerable and measurable, but two findings caution against reading them as evidence of construct-valid personality. First, LLM self-reports are unstable: semantically equivalent prompts and reordered response options produce significantly different trait scores, so the measurement itself is unreliable \citep{gupta2024}. Second, models display a human-like social-desirability bias, skewing toward desirable trait poles once they infer that a personality assessment is underway \citep{salecha2024} — the same evaluative halo we must guard against in our own criterion analysis (\S\ref{sec:criterion}). Our approach departs from this line in two ways: we replace the free-text persona with a \textit{trained, continuous latent} that is fit to human factor scores rather than described in words, and we validate at the level of population factor structure and nomological network rather than mean trait recovery, sidestepping the prompt-sensitivity failure mode by never phrasing the trait as a natural-language instruction.

\subsection{Construct validity as the evaluation bar}

\deleted{We hold the generation of synthetic population to the psychometric notion of}\added{We hold synthetic populations to the psychometric standard of} \textit{construct validity}: a construct is defined not by any single indicator but by its place in a \textit{nomological network} of relationships to other measures \citep{cronbachmeehl1955}. Reproducing a construct therefore means reproducing (i) its internal structure — the factor pattern and inter-item covariance, for which we use Tucker's congruence coefficient \citep{tucker1951,lorenzoseva2006} and Cronbach's $\alpha$ \citep{cronbach1951} — and (ii) its external relationships, for which we use \deleted{meta-analytically documented}\added{published} Big Five correlates of grit \citep{duckworth2009}, self-esteem \citep{rosenberg1965,robins2001}, leadership emergence \citep{judge2002}, and procrastination \citep{steel2007} as criterion targets. This is a strictly more demanding bar than the mean-trait-recovery criterion of the personality-prompting literature: a model can shift its trait scores correctly while failing to reproduce either the factor structure or the nomological links, and it is precisely those latter properties that determine whether a synthetic sample is usable for factor-analytic scale development.

\subsection{Conditioning mechanisms}

Injecting a dense trait code into a frozen backbone is a parameter-efficient fine-tuning problem. We combine soft prefixes / prompt tuning \citep{lester2021} with low-rank adapters (LoRA) \citep{hu2021} and read answers generatively off the digit logits, keeping the language head intact so the same model can also produce free text. This is a distinct design point from inference-time activation steering, in which a trait or behavior direction is added to hidden states without training a mapping from a continuous construct latent to responses \citep{zou2023}. Steering shows that trait-like directions exist in a model's representation space; our aim is different — to \textit{learn} a construct$\rightarrow$response mapping whose synthetic outputs reproduce human psychometric structure — and the soft-prefix-plus-LoRA path lets us fit that mapping from human factor scores while leaving the readout generative.

\section{Method}
\label{sec:method}

\subsection{Overview and notation}
\label{sec:overview}

\begin{figure}[t!]
\centering
\includegraphics[width=0.92\linewidth]{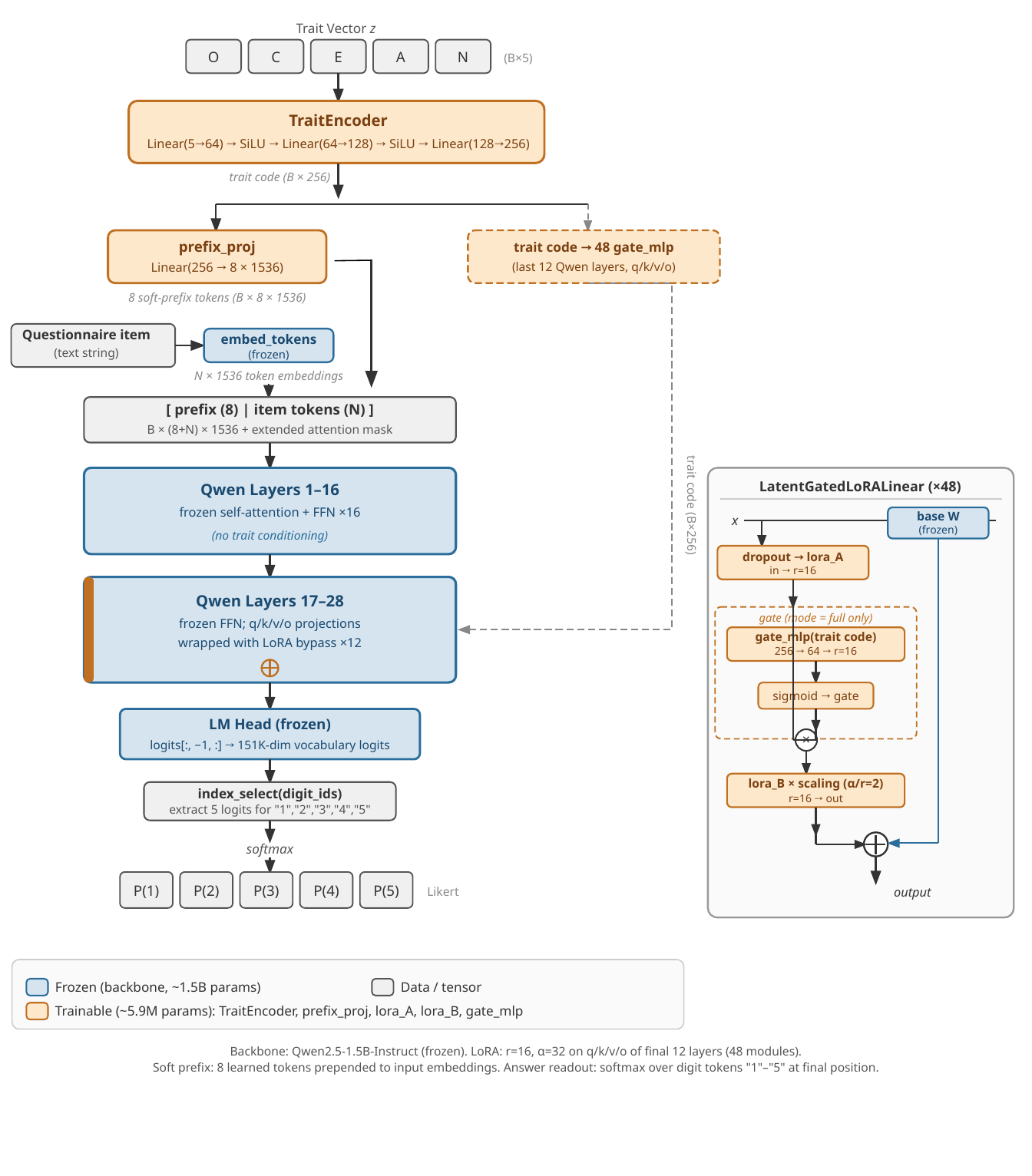}
\caption{Architecture of the respondent simulator (shown for the Qwen2.5-1.5B backbone). Blue components belong to the frozen backbone ($\approx$1.5B parameters, including the token embeddings and the language-modeling head); orange components are the trainable conditioning modules ($\approx$5.9M parameters): the trait encoder $E_\phi$ (Equation~\ref{eq:traitcode}), the prefix projection, and the latent-gated low-rank adapters wrapped around the query/key/value/output projections of the final 12 transformer layers. The five-dimensional latent $\mathbf{z}$ is encoded into a 256-dimensional trait code that (i) produces $m = 8$ soft prefix embeddings prepended to the item's token embeddings and (ii) drives the per-rank sigmoid gates of the 48 adapter modules (inset: a single wrapped projection, Equation~\ref{eq:lora}). The Likert response distribution is read from the final-position next-token logits restricted to the digit tokens ``1''--``5'' (Equation~\ref{eq:readout}). The dashed gate path is active only in the \textit{full} condition; the ungated ablation fixes $\mathbf{g} = \mathbf{1}$ (Section~\ref{sec:baselines}).}
\label{fig:architecture}
\end{figure}

We model a questionnaire respondent as a conditional distribution over Likert responses. Let $\mathbf{z} = (z_O, z_C, z_E, z_A, z_N)^\top \in \mathbb{R}^5$ denote a respondent's standing on the five personality factors, expressed as standardized factor scores; let $x_j$ denote the text of item $j$; and let $Y_{ij} \in \{1, \dots, K\}$ denote the Likert response of respondent $i$ to item $j$ (the training inventory uses $K = 5$). The simulator defines a conditional categorical response distribution
\begin{equation}
p_\theta\!\left(Y_{ij} = k \mid x_j, \mathbf{z}_i\right), \qquad k = 1, \dots, K,
\label{eq:model}
\end{equation}
parameterized by a frozen, instruction-tuned language model augmented with a small set of trainable conditioning modules, collectively denoted $\theta$. A synthetic respondent population --- a \textit{silicon sample} --- is then obtained in two steps: latent vectors $\mathbf{z}_i$ are drawn from the empirical distribution of human factor scores, and each synthetic respondent's item responses are sampled from Equation~\ref{eq:model} (Section~\ref{sec:silicon}). Figure~\ref{fig:architecture} gives an overview of the architecture; the following subsections define each component.

\subsection{Backbone and generative Likert readout}
\label{sec:readout}

The backbone is a frozen, instruction-tuned, causal language model. \deleted{We used two relatively small Llms, that can run on a commercial GPU, in our case a Nvidia RTX 4090 with 24gb of memory:  Qwen2.5-1.5B-Instruct; Llama-3-8B-Instruct (4-bit quantized) .}\added{We use two comparatively small open-weight models that fit on a single consumer GPU (NVIDIA RTX~4090, 24\,GB): Qwen2.5-1.5B-Instruct \citep{qwen2025} and Llama-3-8B-Instruct \citep{llama3_2024}, the latter loaded with 4-bit weight quantization and labelled Llama-3-8B-4bit in the tables.} All backbone parameters --- token embeddings, transformer layers, and the language-modeling head --- remain frozen throughout training and evaluation.

Each item is presented in the backbone's chat format with the system message ``You are a psychometric respondent simulator.'' and a user turn containing the item text followed by the instruction ``On a scale of 1 to 5 (1 = Strongly Disagree, 5 = Strongly Agree), how much do you agree? Answer with a single number.'' Let $\boldsymbol{\ell} \in \mathbb{R}^{|\mathcal{V}|}$ denote the next-token logits produced by the frozen language-modeling head at the final position of this prompt, where $\mathcal{V}$ is the vocabulary, and let $t_1, \dots, t_5$ index the vocabulary entries of the digit tokens ``1'' through ``5''. The response distribution is the temperature-scaled softmax restricted to these five tokens:
\begin{equation}
p_\theta\!\left(Y = k \mid x, \mathbf{z}\right) \;=\; \frac{\exp\!\left(\ell_{t_k} / T\right)}{\sum_{k'=1}^{5} \exp\!\left(\ell_{t_{k'}} / T\right)},
\label{eq:readout}
\end{equation}
where $T > 0$ is a sampling temperature ($T = 1$ during training; its role at generation time is examined in \S\ref{sec:temperature}). This generative readout was chosen over a dedicated classification head for two reasons. First, because the answer is produced by the model's own language-modeling head rather than a dedicated classifier, the language head is left intact, so in principle the same model can still produce open-ended text under the same trait conditioning (a capability we note but do not evaluate in this paper). Second, Equation~\ref{eq:readout} yields a proper categorical distribution for every (item, respondent) pair, which can be \textit{sampled}; Section~\ref{sec:silicon} explains why sampling, rather than taking the modal response, is essential for reproducing psychometric structure.

\subsection{Latent conditioning}
\label{sec:conditioning}

The latent $\mathbf{z}$ is injected into the frozen backbone through two trainable, parameter-efficient paths that share a common representation. A trait encoder $E_\phi$ --- a three-layer perceptron ($5 \to 64 \to 128 \to 256$, SiLU activations) --- first maps the latent to a trait code
\begin{equation}
\mathbf{c} \;=\; E_\phi(\mathbf{z}) \;\in\; \mathbb{R}^{256}.
\label{eq:traitcode}
\end{equation}

\paragraph{Path 1: soft prefix.} An affine projection maps the trait code to $m = 8$ continuous prefix embeddings,
\begin{equation}
\mathbf{P} \;=\; \operatorname{reshape}\!\left(\mathbf{W}_{\!P}\,\mathbf{c} + \mathbf{b}_P\right) \;\in\; \mathbb{R}^{m \times d},
\label{eq:prefix}
\end{equation}
which are prepended to the item's token-embedding sequence, $[\,\mathbf{P};\, \mathbf{E}_{\text{emb}}(x_j)\,]$, with the attention mask extended accordingly \citep{lester2021}. The projection weights are initialized with small variance ($\mathcal{N}(0, 0.02^2)$, zero bias), so at the start of training the prefix is close to a null input and the conditioned model's behavior starts near that of the unconditioned backbone.

\paragraph{Path 2: latent-gated low-rank adaptation.} Each of the query, key, value, and output projections in the self-attention blocks of the final 12 transformer layers (of 28 for Qwen2.5-1.5B; of 32 for Llama-3-8B) --- 48 modules in total --- is wrapped with a low-rank adapter \citep{hu2021} of rank $r = 16$ and scaling $\alpha = 32$. Writing $\mathbf{W}_0$ for the frozen projection weight and $\mathbf{u}$ for its input, the wrapped projection computes
\begin{equation}
\mathbf{h} \;=\; \mathbf{W}_0 \mathbf{u} \;+\; \frac{\alpha}{r}\, \mathbf{B}\left(\mathbf{g} \odot \mathbf{A}\tilde{\mathbf{u}}\right), \qquad \mathbf{g} \;=\; \sigma\!\left(G_\psi(\mathbf{c})\right) \in (0,1)^r,
\label{eq:lora}
\end{equation}
where $\mathbf{A} \in \mathbb{R}^{r \times d_{\text{in}}}$ and $\mathbf{B} \in \mathbb{R}^{d_{\text{out}} \times r}$ are trainable ($\mathbf{B}$ is zero-initialized, so the bypass initially contributes nothing), $\tilde{\mathbf{u}}$ is the input after dropout ($p = .05$), $G_\psi$ is a two-layer gate network ($256 \to 64 \to r$, SiLU), $\sigma$ is the logistic function, and $\odot$ denotes elementwise multiplication. The gate $\mathbf{g}$ modulates each of the $r$ bottleneck dimensions as a function of the trait code, allowing the adapter's effective update to vary with the respondent's trait profile. Fixing $\mathbf{g} = \mathbf{1}$ recovers standard (ungated) LoRA; this ungated variant is evaluated as an ablation, and --- anticipating \S\ref{sec:ablation} --- the gate confers no reliable advantage, so we do not claim it as a contribution of the method.

Across the trait encoder, prefix projection, adapters, and gate networks, approximately 5.9M parameters are trainable for the Qwen backbone --- roughly 0.4\% of the backbone's parameter count. For the larger Llama-3-8B backbone the same modules total approximately 14.4M trainable parameters ($\approx$0.18\% of the backbone), the increase being driven by its wider hidden size (4096 vs.\ 1536). Everything else is frozen.

\subsection{Training data and objective}
\label{sec:training}

Training data are item-level responses from the IPIP-50 Big Five inventory ($N = 15{,}775$ participants\added{; source: the BIG5 dataset of the Open-Source Psychometrics Project\footnote{\url{https://openpsychometrics.org/_rawdata/} (archive \texttt{BIG5.zip}).}, collected c.~2012 through an online administration of the IPIP Big-Five Factor Markers \citep{goldberg1992} and comprising 19,719 respondents; the remaining 3,944 participants form the participant holdout of Section~\ref{sec:instruments}}). Each participant's latent $\mathbf{z}_i$ is their vector of factor scores from a confirmatory five-factor model fitted to the human data (Section~\ref{sec:instruments} describes the item partitions; 30 of the 50 items are used for training and 20 --- four per factor --- are withheld from the training objective). The confirmatory model that produces $\mathbf{z}$ is itself fitted to all 50 IPIP items\added{ in the training participants}, so each latent reflects the full inventory; the held-out items' \textit{individual responses}, however, never enter the training loss, which scores only the 30 training items. With $\mathcal{D}$ the set of observed (participant, item) response pairs over the training items, the trainable modules minimize the negative log-likelihood of the observed responses under Equation~\ref{eq:readout} at $T = 1$:
\begin{equation}
\mathcal{L}(\theta) \;=\; -\,\frac{1}{|\mathcal{D}|} \sum_{(i,j) \in \mathcal{D}} \log p_\theta\!\left(Y = y_{ij} \mid x_j,\, \mathbf{z}_i + \boldsymbol{\varepsilon}_{ij}\right), \qquad \boldsymbol{\varepsilon}_{ij} \sim \mathcal{N}\!\left(\mathbf{0},\, \sigma_\varepsilon^2 \mathbf{I}_5\right),
\label{eq:loss}
\end{equation}
where the isotropic jitter $\boldsymbol{\varepsilon}_{ij}$ ($\sigma_\varepsilon = 0.1$, redrawn at every presentation) regularizes the conditioning channel against overfitting to exact factor-score values; validation uses unperturbed latents. Optimization uses AdamW (learning rate $2 \times 10^{-4}$, weight decay 0.01), batch size 16, three epochs, and a schedule of 3\% linear warmup followed by cosine decay, in bfloat16 precision. Model selection is by cross-entropy on responses to the 20 held-out items, so the selected checkpoint is the one that best generalizes beyond the items seen in the loss rather than the one that best fits the training items. Because selection is by pointwise per-item cross-entropy whereas the reported structural metric is factor congruence (a second-order property of the item correlation matrix), the resulting selection bias on the \S\ref{sec:heldout} held-out estimate \deleted{is minimal}\added{is expected to be small}; moreover, the cross-inventory \added{results} ({\S\ref{sec:bfi}})\deleted{ and criterion ({\S\ref{sec:criterion}}) results, which do not inform checkpoint selection at all,}\added{, which share no items with the selection set,} corroborate that estimate. \added{The criterion analysis (\S\ref{sec:criterion}) is less independent in this respect, because its observed Big Five scores are computed from the same 20 items used for selection. Five independent seeds were trained for each backbone; unless stated otherwise, all aggregates are mean $\pm$ SD over the five seeds.}

\subsection{Constructing a silicon sample}
\label{sec:silicon}

A silicon sample of $n$ synthetic respondents is generated in two steps.

\begin{enumerate}
\item \textbf{Population latents.} Respondent latents are drawn from a multivariate normal matched to the human factor-score distribution,
\begin{equation}
\mathbf{z}_i \;\overset{\text{iid}}{\sim}\; \mathcal{N}_5\!\left(\boldsymbol{\mu}_{\text{human}},\, \Sigma_{\text{human}}\right), \qquad i = 1, \dots, n,
\label{eq:latents}
\end{equation}
where $\boldsymbol{\mu}_{\text{human}}$ and $\Sigma_{\text{human}}$ are the empirical mean vector and covariance matrix of the human CFA factor scores. Because $\Sigma_{\text{human}}$ encodes the human inter-factor covariance, the synthetic population reproduces the inter-factor correlation structure $\Phi$ \textit{at the latent level} by construction; whether that structure survives the mapping into item responses is an empirical question (\S\ref{sec:heldout}).
\item \textbf{Response sampling.} Each synthetic respondent answers every administered item by sampling from the model's categorical at temperature $T$,
\begin{equation}
Y_{ij} \;\sim\; \operatorname{Categorical}\!\left(p_\theta(\,\cdot \mid x_j, \mathbf{z}_i)\right).
\label{eq:sampling}
\end{equation}
\end{enumerate}

Sampling, rather than taking the modal response $\arg\max_k p_\theta(k \mid x_j, \mathbf{z}_i)$, is essential: the argmax collapses each respondent--item distribution to a deterministic function of $\mathbf{z}$, discarding the item-unique variance on which factor analysis operates. Unless stated otherwise, the administered items are the \textit{held-out} items, so all structural results concern items the model never saw during training.

\subsection{Baseline and ablation conditions}
\label{sec:baselines}

Four conditioning modes form a ladder from no trait information to the full trained method:

\begin{enumerate}
\item \textit{baseline\_a} (floor): the frozen backbone answers each item with no trait information of any kind.
\item \textit{baseline\_b} (persona prompt): the frozen backbone receives the trait profile as text in the prompt --- either the numeric $z$-scores or a verbal rendering into ordinal labels --- and is instructed to answer as that person would. No weights are trained. This operationalizes the prompt-persona approach of prior work \citep{argyle2023,serapiogarcia2023,jiang2023}; the exact prompt formats are given in \S\ref{sec:ablation}.
\item \textit{baseline\_c} (trained, ungated): the soft prefix and adapters of Section~\ref{sec:conditioning} are trained with the gate disabled ($\mathbf{g} = \mathbf{1}$ in Equation~\ref{eq:lora}).
\item \textit{full} (trained, gated): identical to \textit{baseline\_c} plus the trained gate networks $G_\psi$.
\end{enumerate}

The two trained conditions share the training protocol of Section~\ref{sec:training}. Each trained checkpoint is always evaluated under the conditioning mode it was trained with.

\section{Evaluation methodology}
\label{sec:evaluation}

\subsection{Instruments and data partitions}
\label{sec:instruments}

Three evaluation sets probe increasingly distant generalization from the training data.

\paragraph{IPIP held-out items.} The 50 IPIP items were partitioned at the item level, stratified by factor: 30 items for training and 20 never-trained items (four per factor) constituting the new-item test. Throughout, we use held-out and new-item to mean never present in the training objective, i.e. the model's weights are never updated on these items; note that the 20 held-out items are additionally used for checkpoint selection (Section~\ref{sec:training}), so they are held out from the training loss but not from model selection. A separate participant-holdout set (the 30 training items answered by 3,944 additional participants not used in training) supports a well-powered structural check on seen items.

\paragraph{BFI (out-of-distribution inventory).} The transfer test used the Big Five Inventory--2 \citep[BFI-2;][]{soto2017}, a 60-item inventory (12 items per domain) developed independently of the IPIP. Its items use a distinct format---the stem ``I am someone who\ldots'' followed by short descriptive phrases (e.g., ``is outgoing, sociable''; ``tends to be quiet'')---and none of its items appears in the training inventory, so it tests transfer to an independent instrument rather than to new items from the same pool. Human responses came from a publicly available sample of 1,733 respondents who answered on the same 5-point scale used in training, distributed as a Czech standardization sample of the BFI-2 with the \textit{ShinyItemAnalysis} R package \citep{martinkova2018}. Items were presented to the simulator in their native first-person English format; because the human respondents answered a Czech translation, the transfer test is additionally cross-lingual. The distributed human data are stored already reverse-keyed (higher = more of the trait for every item), whereas the simulator answers the raw content of each item; we therefore restored the reverse-keyed human items to raw form so that human and synthetic responses were scored by the identical pipeline.

\paragraph{Criterion battery.} A four-instrument battery from outside the personality domain (grit, self-esteem, leadership emergence, procrastination), used for the nomological-validity analyses, is described together with its analysis in \S\ref{sec:criterion}.

\subsection{Psychometric indices}
\label{sec:indices}

Human and silicon response matrices are analyzed identically in R (packages \textit{psych} and \textit{GPArotation}): a five-factor exploratory factor analysis (minimum-residual extraction, oblimin rotation) is fitted separately to each matrix, and the resulting solutions are compared on four indices.

\paragraph{Factor congruence.} Let $\boldsymbol{\lambda}^{H}$ and $\boldsymbol{\lambda}^{S}$ denote the loading vectors of a matched human and silicon factor over the $p$ common items. Factor similarity is quantified by Tucker's congruence coefficient \citep{tucker1951},
\begin{equation}
\varphi\!\left(\boldsymbol{\lambda}^{H}, \boldsymbol{\lambda}^{S}\right) \;=\; \frac{\sum_{j=1}^{p} \lambda^{H}_{j}\, \lambda^{S}_{j}}{\sqrt{\left(\sum_{j=1}^{p} \bigl(\lambda^{H}_{j}\bigr)^{2}\right)\left(\sum_{j=1}^{p} \bigl(\lambda^{S}_{j}\bigr)^{2}\right)}},
\label{eq:tucker}
\end{equation}
with factors matched one-to-one by a greedy assignment on the congruence matrix. The mean of Equation~\ref{eq:tucker} over the five matched factors is the headline structural index; values of .85--.94 are conventionally read as fair similarity and $\geq$ .95 as factor equality \citep{lorenzoseva2006}.

\paragraph{Inter-factor structure.} The oblimin factor-correlation matrices $\Phi^{H}$ and $\Phi^{S}$ are compared by the Frobenius norm of their difference, $\lVert \Phi^{S} - \Phi^{H} \rVert_F$ (lower is better).

\paragraph{Internal consistency and item behavior.} For each trait scale of $k$ items with item variances $\sigma^2_j$ and total-score variance $\sigma^2_X$, Cronbach's alpha \citep{cronbach1951},
\begin{equation}
\alpha \;=\; \frac{k}{k-1}\left(1 - \frac{\sum_{j=1}^{k} \sigma^2_j}{\sigma^2_X}\right),
\label{eq:alpha}
\end{equation}
is computed for human and silicon data and compared as $\lvert \alpha^{S} - \alpha^{H} \rvert$, alongside corrected item--total correlations (each item against the sum of the remaining items of its scale).

The full pipeline was validated by splitting the human sample into two disjoint random halves and comparing them with the identical factor-analysis machinery. Because the halves are independent subsamples of one population, this establishes a human noise ceiling for each metric: mean Tucker congruence 0.988 (IPIP, 20 held-out items) and 0.984 (BFI-2, 60 items), with a $\Phi$ Frobenius error of 0.200 and 0.170 respectively and per-scale alphas agreeing to within 0.04. These are the strongest values any silicon population could realistically attain given ordinary sampling variability, so a silicon result approaching this ceiling indicates the simulator rather than the analysis pipeline is the limiting factor; conversely, a residual $\Phi$ error of $\approx 0.2$ is expected even between two groups of humans and should be read as the baseline against which silicon $\Phi$ errors are judged.

\section{Results}
\label{sec:results}

\subsection{Factor-structure reproduction (held-out items)}
\label{sec:heldout}

Table~\ref{tab:ipip-heldout} reports IPIP held-out performance (20 items, 4 per factor) aggregated by backbone. \deleted{Qwen-2.5-1.5B}\added{Qwen2.5-1.5B} achieves a mean Tucker congruence of 0.850 $\pm$ 0.014. Replication is tightest for Conscientiousness (0.90 $\pm$ 0.02) and Agreeableness (0.88 $\pm$ 0.01) and loosest for Neuroticism (0.74 $\pm$ 0.05) and Extraversion (0.84 $\pm$ 0.03). Llama-3-8B-4bit reaches 0.873 $\pm$ 0.034, with particularly high Extraversion (0.94 $\pm$ 0.02) and Conscientiousness (0.91 $\pm$ 0.03); Openness and Neuroticism show wider cross-seed variance (SDs 0.06 and 0.08 respectively). The $\Phi$ Frobenius error and mean corrected item-total correlation are comparable across backbones, and the silicon item-total correlation (0.481 Qwen, 0.494 Llama) closely matches the human value of 0.484 (Table~\ref{tab:ipip-heldout}).

\begin{table}[ht!]
\centering
\caption{Tucker congruence coefficients and related psychometric indices for the IPIP held-out set (20 items, 4 per factor). Tucker: mean Tucker congruence across the five matched factors. O–N: per-factor Tucker congruence for Openness (O), Conscientiousness (C), Extraversion (E), Agreeableness (A), and Neuroticism (N). $\Phi$ error: Frobenius norm of the difference between the silicon and human inter-factor correlation matrices. item-r: mean corrected item-total correlation over the held-out items (human reference 0.484). All values are mean $\pm$ SD across seeds \deleted{(Qwen: four seeds, excluding the bad-init seed 3; Llama: four seeds)}\added{(five seeds per backbone)}. SD reflects cross-seed variability in how consistently each factor is recovered.\added{ Silicon responses were sampled at $T = 1.0$ for both backbones.}}
\label{tab:ipip-heldout}
\resizebox{\textwidth}{!}{%
\begin{tabular}{lcccccccc}
\toprule
Group & Tucker & O & C & E & A & N & $\Phi$ error & item-r \\
\midrule
Qwen  & 0.850 $\pm$ 0.014 & 0.90 $\pm$ 0.01 & 0.90 $\pm$ 0.02 & 0.84 $\pm$ 0.03 & 0.88 $\pm$ 0.01 & 0.74 $\pm$ 0.05 & 0.640 $\pm$ 0.060 & 0.481 $\pm$ 0.024 \\
Llama  & 0.873 $\pm$ 0.034 & 0.89 $\pm$ 0.06 & 0.91 $\pm$ 0.03 & 0.94 $\pm$ 0.02 & 0.87 $\pm$ 0.03 & 0.76 $\pm$ 0.08 & 0.729 $\pm$ 0.103 & 0.494 $\pm$ 0.028 \\
\bottomrule
\end{tabular}
}
\end{table}

\deleted{The inter-factor correlation matrix is well-preserved: silicon scale-score correlations correlate r = 0.95 with the human factor-score covariance (E–A positive, C–N negative, E–N negative all reproduced).}\added{The pattern of inter-factor correlations is preserved: the silicon inter-scale correlations correlate $r = 0.95$ with the corresponding human inter-factor correlations (E--A positive, C--N negative, E--N negative all reproduced), which is partly expected because $\Sigma_{\text{human}}$ is injected at the latent level (Equation~\ref{eq:latents}). Absolute values are reproduced less faithfully: the $\Phi$ error (0.64--0.73) is roughly three times the human split-half ceiling of 0.20 (Section~\ref{sec:indices}).} On a well-powered structural check — 30 items, 6 per factor, seen items, new synthetic respondents — Tucker congruence reaches 0.986 with all five factors at 0.98–0.99 and $\Phi$ error 0.224\deleted{, confirming that the lower held-out figures reflect 4-items-per-factor FA sampling noise rather than a structural failure of the conditioning}. \added{Because this check differs from the held-out evaluation both in items per factor and in training exposure, it cannot separate a short-scale effect from a generalization gap; and since two human half-samples reach 0.988 on the same 20 held-out items (Section~\ref{sec:indices}), short scales alone cannot explain the shortfall. We therefore read the gap between 0.85--0.87 and the 0.99 ceiling as a genuine, moderate loss of structure on unseen items, concentrated in Neuroticism.}

\subsection{Cross-inventory transfer (BFI-2, out of distribution)}
\label{sec:bfi}

Table~\ref{tab:bfi-ood} reports performance on the 60 BFI-2 items (different inventory, different wording) — items the model never saw during training. Qwen achieves a mean Tucker congruence of 0.806 $\pm$ 0.007, with the strongest replication on Openness (0.85 $\pm$ 0.02) and Conscientiousness (0.84 $\pm$ 0.02); Neuroticism is again the weakest factor (0.74 $\pm$ 0.05). Llama reaches 0.803 $\pm$ 0.074.

\deleted{Taken together, both backbones transfer to a completely different instrument with different item wording, which can only arise if the model has learned construct-to-response behaviour rather than surface item features. The $\Phi$ Frobenius error and item-total correlations are comparable to the IPIP held-out figures (Table~{\ref{tab:bfi-ood}} vs Table~{\ref{tab:ipip-heldout}}). Anyway, we analyze in section {\S\ref{sec:temperature}} the role of Temperature and found that Tucker congruence is nonetheless robust to temperature choice.}\added{Both backbones thus transfer to an independently developed instrument with different wording and stem format and, because the human sample answered a Czech translation, across languages. This is difficult to explain by surface item features and is most naturally read as construct-to-response behaviour, although the two inventories necessarily share content at the level of the trait domains. The $\Phi$ error is comparable to the IPIP held-out figure, whereas silicon corrected item--total correlations (0.41--0.42) fall below the human reference (0.50): silicon BFI-2 items are somewhat noisier indicators of their scales. Section~\ref{sec:temperature} shows that these congruence values are robust to the sampling temperature.}

\begin{table}[ht!]
\centering
\caption{Tucker congruence coefficients and related psychometric indices for the BFI-2 out-of-distribution set (60 items, 12 per factor). Column definitions as in Table~\ref{tab:ipip-heldout} (item-r human reference 0.500). Values are mean $\pm$ SD across seeds \deleted{(Qwen: four, excluding seed 3; Llama: four)}\added{(five seeds per backbone), at the same sampling temperatures as Table~\ref{tab:ipip-heldout}}.}
\label{tab:bfi-ood}
\resizebox{\textwidth}{!}{%
\begin{tabular}{lcccccccc}
\toprule
Group & Tucker & O & C & E & A & N & $\Phi$ error & item-r \\
\midrule
Qwen  & 0.806  $\pm$ 0.007 & 0.85 $\pm$ 0.02 & 0.84 $\pm$ 0.02 & 0.81 $\pm$ 0.01 & 0.80 $\pm$ 0.02 & 0.74 $\pm$ 0.05 & 0.610 $\pm$ 0.040 & 0.409 $\pm$ 0.011 \\
Llama  & 0.803 $\pm$ 0.074 & 0.83 $\pm$ 0.11 & 0.80 $\pm$ 0.05 & 0.80 $\pm$ 0.12 & 0.84 $\pm$ 0.05 & 0.76 $\pm$ 0.08 & 0.752 $\pm$ 0.111 & 0.419 $\pm$ 0.037 \\
\bottomrule
\end{tabular}
}
\end{table}

\subsection{Baselines and ablation ladder}
\label{sec:ablation}

We evaluate four conditioning modes that form a ladder from no trait information to the full trained method. All modes share the same system prompt (``You are a psychometric respondent simulator.'') and the same Likert instruction appended to the user turn (``On a scale of 1 to 5 (1 = Strongly Disagree, 5 = Strongly Agree), how much do you agree? Answer with a single number.''). They differ in whether and how the Big Five latent z = (O, C, E, A, N) is communicated to the model.

Baseline\_a (floor) uses the frozen backbone with no trait information — the item text is the sole content of the user turn. No weights are updated. This measures what psychometric structure a general instruction model produces when answering personality items without any conditioning.

Baseline\_b (persona prompt) keeps the backbone frozen and injects the trait profile as text in the user turn. Two formats are tested. In the numeric style the latent z-scores are prepended as a header block:

\begin{verbatim}
Participant Profile
Openness: 0.72
Conscientiousness: -0.34
Extraversion: 1.05
Agreeableness: 0.18
Neuroticism: -0.82

Question: [item text]
On a scale of 1 to 5 ...
\end{verbatim}

In the verbal style the z-scores are first mapped to ordinal labels ($z \geq 1.0 \rightarrow$ ``very high''; $0.5 \leq z < 1.0 \rightarrow$ ``high''; $-0.5 < z < 0.5 \rightarrow$ ``average''; $-1.0 < z \leq -0.5 \rightarrow$ ``low''; $z \leq -1.0 \rightarrow$ ``very low'') and the prompt opens with a role-play instruction:

\begin{verbatim}
You are role-playing a specific person with this personality profile (Big Five):
- Openness: high
- Conscientiousness: average
- Extraversion: very high
- Agreeableness: average
- Neuroticism: low

Answer the question as this person would.

Question: [item text]
On a scale of 1 to 5 ...
\end{verbatim}

No weights are updated in either variant; the model relies entirely on its instruction-following prior to interpret the profile.

Baseline\_c (trained, no gate) fine-tunes the soft prefix and LoRA adapters described in Section~\ref{sec:conditioning}, with the gate MLP disabled. The text prompt reverts to the baseline\_a format — no numeric or verbal profile appears in the prompt — because the trait information is delivered entirely through the 8 continuous prefix tokens prepended to the input embeddings. Three training epochs, item-holdout model selection.

Full (trained, gated LoRA) is identical to baseline\_c with the addition of the gate MLP that modulates the LoRA bottleneck from the trait code (Section~\ref{sec:conditioning}). Same training protocol.

Table~\ref{tab:ablation} reports mean Tucker congruence for each mode on the IPIP held-out items and the BFI-2 out-of-distribution inventory, all for the Qwen2.5-1.5B backbone (seed 0). Values for baseline\_a and baseline\_b are zero-shot (no trained weights, hence no seed variance). On the IPIP held-out items trained latent conditioning (0.84–0.86) nearly doubles the best persona prompt (0.46–0.50); on the BFI-2 OOD inventory the margin is smaller but still substantial \added{(0.79–0.80 vs 0.61–0.63)}. Both trained and persona conditioning sit far above the no-trait floor (0.23–0.28). Verbal and numeric persona styles perform similarly, indicating the gap between prompting and trained conditioning is not an artefact of prompt format.

\begin{table}[ht!]
\centering
\caption{Mean Tucker congruence for each conditioning mode for the Qwen2.5-1.5B backbone (seed 0) on the IPIP held-out set (20 items) and the BFI-2 out-of-distribution inventory (60 items). All modes use the frozen backbone; baseline\_a and baseline\_b update no weights. \added{The final row is an architecture ablation that removes the language model entirely (Embedding-MLP; mean over five seeds).}}
\label{tab:ablation}
\begin{tabular}{lcc}
\toprule
Mode & IPIP held-out  & BFI-2 OOD  \\
\midrule
Baseline\_a — frozen LM, no traits & 0.284 & 0.228 \\
Baseline\_b — persona prompt, verbal & 0.464 & 0.606 \\
Baseline\_b — persona prompt, numeric & 0.496 & 0.630 \\
Baseline\_c — trained prefix + LoRA, no gate & 0.856 & \added{0.786} \\
Full — trained prefix + gated LoRA & 0.842 & 0.802 \\
\midrule
\added{Embedding-MLP — mpnet item embedding + $z$, no LLM} & \added{0.402} & \added{0.281} \\
\bottomrule
\end{tabular}
\end{table}

\begin{figure}[ht!]
\centering
\includegraphics[width=\linewidth]{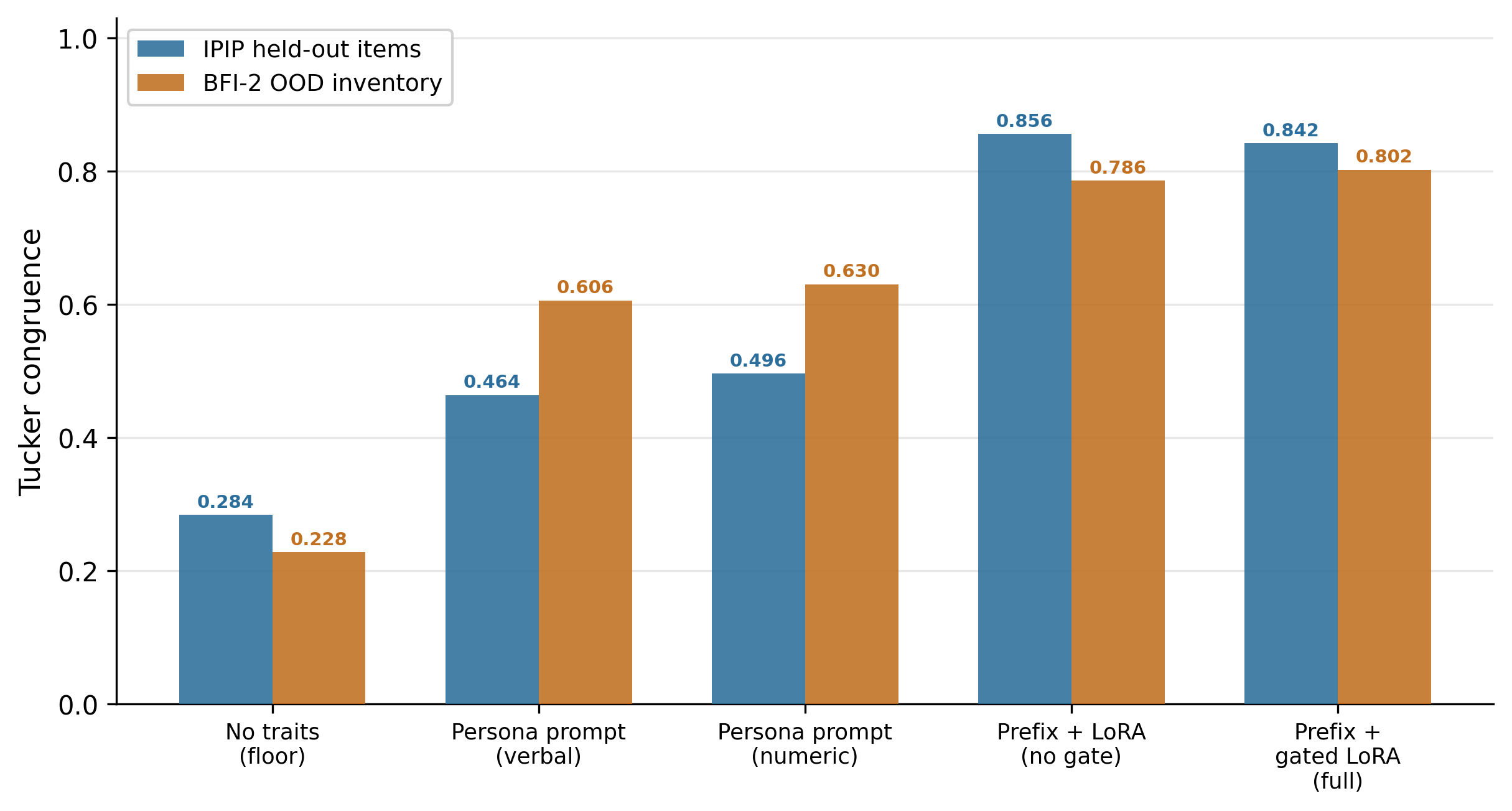}
\caption{Ablation ladder: Tucker factor congruence for each conditioning mode (Qwen2.5-1.5B, seed 0)
on IPIP held-out items (blue) and the BFI-2 out-of-distribution inventory (orange).
Trained latent conditioning \added{(the two rightmost groups, prefix~+~LoRA and prefix~+~gated LoRA)} exceeds persona-prompting, which in turn exceeds the
no-trait floor. The
latent gate adds negligible benefit. All values \added{match Table~\ref{tab:ablation} and are} from matched 3-epoch runs.}
\label{fig:ablation}
\end{figure}

\deleted{From table 3 we can see that Baseline\_c and the full model are basically equivalent, we therefore}\added{Table~\ref{tab:ablation} shows that baseline\_c and the full model are practically equivalent: in this single-seed comparison the gate lowers IPIP congruence by 0.014 and raises BFI-2 congruence by 0.016. We therefore} do not claim latent gating as a contribution; the contributions are the prefix+LoRA continuous-latent conditioning, the sampling-based silicon-sample construction, and the validity evidence. \deleted{We ship \texttt{full} as an operational choice (slightly better, more balanced OOD); baseline\_c is a near-equivalent simpler fallback.}\added{We report \texttt{full} throughout for continuity with the main experiments; baseline\_c is an equally defensible, simpler alternative with $\approx$0.84M fewer trainable parameters.}

\added{A final ablation asks whether the pretrained backbone is necessary at all, or whether the item text together with the latent already suffices. We replaced the language model with a small multilayer perceptron that maps a sentence embedding of the item text (all-mpnet-base-v2, 768 dimensions, mean-pooled and $L_2$-normalised) concatenated with the latent $z$ to the five Likert logits; it was trained on the identical 30 training items, factor scores, and raw responses --- with the same latent jitter and item-holdout model selection as the simulator --- and its synthetic population was built and scored through the identical sampling and factor-analytic pipeline (seeds 0--4). The embedding-MLP \deleted{collapses to roughly the no-trait floor}\added{falls far below the trained simulator and below even the zero-shot persona prompt, approaching the no-trait floor on the BFI-2}: mean Tucker congruence $0.402 \pm 0.020$ on the IPIP held-out items and $0.281 \pm 0.028$ on the BFI-2 (last row of Table~\ref{tab:ablation})\deleted{, below even the zero-shot persona prompt on both inventories}, with near-zero corrected item-total correlations ($\approx 0.04$--$0.07$ against a human reference of $\approx 0.48$--$0.50$) and a large inter-factor error. The MLP fits the training items (train cross-entropy $1.35$) but its held-out cross-entropy ($1.59$) is essentially chance ($\ln 5 = 1.61$): a generic sentence embedding carries \deleted{no signal}\added{little usable signal} for generalising the construct-to-response mapping to unseen item wording or a new inventory. The frozen language model's pretrained semantics, not the conditioning latent on its own, are what enable the held-out and cross-inventory structure reported above.}

\subsection{Criterion and nomological validity}
\label{sec:criterion}

The structural analyses above show that silicon respondents reproduce human factor structure on held-out items. A complementary question is whether the population also recovers the relationships between personality and non-personality outcomes — what psychometric standards call criterion or nomological network validity. This matters particularly because the training setup carries a residual circularity: the conditioning latents are CFA factor scores from the IPIP data, and the inter-factor covariance $\Sigma_{\text{human}}$ is injected directly. Evidence that the model reproduces trait correlates that are absent from $\Sigma_{\text{human}}$ and from the training items would indicate genuine construct-to-response learning rather than injected covariance replay.

To test this we administered four instruments from outside the personality domain whose Big Five correlates are established in \deleted{the meta-analytic literature}\added{published research (meta-analyses for leadership emergence and procrastination, primary studies for grit and self-esteem)}: the Grit-S scale (8 items; dominant trait +C; \citealp{duckworth2009}), the Rosenberg Self-Esteem Scale (10 items; dominant trait $-$N; \citealp{robins2001}), \deleted{a public-domain leadership emergence measure}\added{a public-domain set of IPIP leadership/assertiveness items \citep{goldberg1999}} (10 items; dominant trait +E; \citealp{judge2002}), and \deleted{a procrastination scale}\added{the Pure Procrastination Scale \citep{steel2010}} (8 items; dominant trait $-$C; \citealp{steel2007}). Twenty held-out IPIP Big Five marker items were also included to produce an observed trait score for each synthetic respondent alongside the injected latent z. The battery was administered to n = 1000 synthetic respondents drawn from $\mathcal{N}(\boldsymbol{\mu}_{\text{human}}, \Sigma_{\text{human}})$ (Equation~\ref{eq:latents}) at T = 1.0; domain scores were correlated against the observed Big Five scale scores. The domain–trait relationships tested here are nowhere in $\Sigma_{\text{human}}$: the training procedure injects only the Big Five inter-factor structure, not any link between personality and grit, self-esteem, leadership, or procrastination.

Table~\ref{tab:criterion-zero} reports the silicon zero-order correlations alongside the \deleted{meta-analytic}\added{literature} targets. All four domains recover the correct sign on their theoretically dominant trait, with bootstrap 95\% CIs excluding zero on the dominant-trait correlation in every case. Profile correlations with the literature vectors range from 0.73 (procrastination) to 0.95 (self-esteem), with a mean of 0.87 (Qwen\added{ seed~0}, n = 1000); the Llama checkpoints reach a comparable mean profile correlation (0.89, \deleted{n = 4}\added{mean over five checkpoints}). The floor condition (baseline\_a, no traits) yields mean profile correlations near zero, confirming that the signal is produced by trait conditioning rather than by item content or the sampling procedure.

\begin{table}[ht!]
\centering
\caption{Silicon zero-order Pearson correlations between each domain score and observed Big Five scale scores (Qwen\added{ seed~0}, n = 1000, T = 1.0), alongside the \deleted{meta-analytic }literature target vector used for validation (sources: \citep{duckworth2009,robins2001,judge2002,steel2007}). Profile r: Pearson correlation between the five-element silicon r-vector and the literature target vector. The theoretically dominant trait is in the expected direction for all four domains; bootstrap 95\% CIs on the dominant-trait correlation exclude zero in every case. The floor (baseline\_a, no traits) gives mean profile r $\approx$ 0.}
\label{tab:criterion-zero}
\resizebox{\textwidth}{!}{%
\begin{tabular}{lccccccc}
\toprule
Domain & Literature target (O / C / E / A / N) & Silicon $r_O$ & Silicon $r_C$ & Silicon $r_E$ & Silicon $r_A$ & Silicon $r_N$ & Profile r \\
\midrule
Grit (+C) & .06 / .77 / .20 / .24 / $-$.40 & +.39 & +.54 & +.21 & +.38 & $-$.36 & .88 \\
Self-esteem ($-$N) & .17 / .24 / .38 / .13 / $-$.50 & +.48 & +.52 & +.44 & +.47 & $-$.50 & .95 \\
Leadership (+E) & .24 / .28 / .31 / .08 / $-$.24 & +.49 & +.38 & +.65 & +.46 & $-$.07 & .91 \\
Procrastination ($-$C) & .03 / $-$.62 / $-$.12 / $-$.12 / .24 & $-$.38 & $-$.52 & $-$.36 & $-$.38 & +.56 & .73 \\
\bottomrule
\end{tabular}
}
\end{table}

A serious concern with the zero-order profile is that an evaluative halo — a shared positive loading on socially desirable traits and a negative one on Neuroticism — could inflate every correlation and manufacture apparent criterion validity without any genuine construct learning. Because the \deleted{meta-analytic }literature target vectors are themselves close to this halo shape, the profile-correlation metric has limited power to separate the two. We make this explicit with two reference baselines computed against the identical literature targets (Table~\ref{tab:criterion-baselines}). A pure halo vector — $(+,+,+,+,-)$ on O/C/E/A/N, oriented a priori by domain valence and carrying no model information — attains a mean profile r of 0.82 versus the model's 0.87, and in fact matches or exceeds the model on Self-esteem and Leadership; it also reproduces the correct dominant-trait sign in all four domains by construction, so the ``correct sign in all cells'' headline carries little discriminating value. The model exceeds the halo only on Grit and Procrastination — the two domains whose literature profile is least halo-shaped (a single strong Conscientiousness loading). A second, semantic baseline predicts each domain's trait profile from item text alone — embedding every battery item with the frozen backbone and taking the cosine between sign-keyed domain and trait directions — and reaches a mean profile r of only 0.35, well below both the model and the halo; item wording therefore does not by itself reproduce the profiles. Taken together, the zero-order profile correlation and the sign test are necessary but not sufficient. We therefore rest the nomological claim on the discriminant $\beta$ analysis below, which a halo cannot satisfy: we regress each domain score on all five Big Five predictors simultaneously (z-scored inputs and outcomes), yielding standardized $\beta$ coefficients that partial out the shared evaluative variance and isolate each factor's unique contribution.

\begin{table}[H]
\centering
\caption{Profile-correlation baselines for the criterion metric, against the identical \deleted{meta-analytic }literature targets. \textbf{Model}: silicon zero-order profile r (Qwen seed 0, n = 1000). \textbf{Halo}: a pure evaluative-halo vector $(+,+,+,+,-)$ oriented by domain valence, carrying no model information. \textbf{Semantic}: profile predicted from item text alone (frozen-backbone embeddings, sign-keyed domain-vs-trait cosine). \textbf{Floor}: no-trait baseline\_a. A pure halo nearly matches the model, exposing the metric's limited discriminating power; item semantics alone do not.}
\label{tab:criterion-baselines}
\begin{tabular}{lcccc}
\toprule
Domain & Model & Halo & Semantic & Floor \\
\midrule
Grit & 0.88 & 0.77 & 0.14 & $-$0.12 \\
Self-esteem & 0.95 & 0.96 & 0.57 & 0.11 \\
Leadership & 0.91 & 0.92 & 0.25 & $-$0.22 \\
Procrastination & 0.73 & 0.63 & 0.44 & 0.24 \\
\midrule
Mean & 0.87 & 0.82 & 0.35 & 0.01 \\
\bottomrule
\end{tabular}
\end{table}

\begin{figure}[ht!]
\centering
\includegraphics[width=\linewidth]{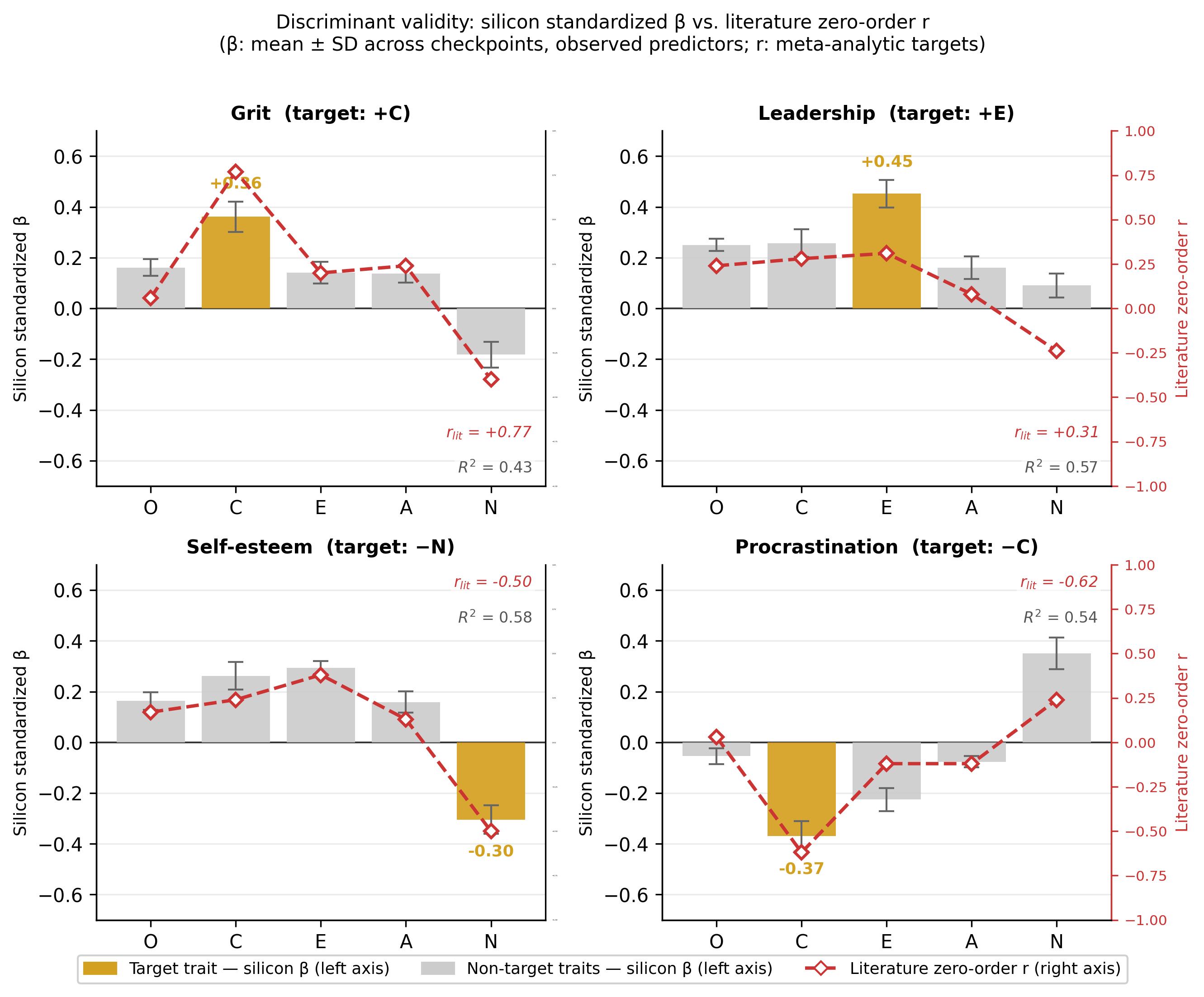}
\caption{Discriminant validity: silicon standardized $\beta$ profiles (bars) vs.\ \deleted{meta-analytic}\added{published}
literature zero-order $r$ values (dashed line) for each criterion domain. Amplitude difference
between bars and line reflects variance partialling (betas control for inter-trait correlations;
zero-order $r$ does not). \deleted{Profile shapes match in every domain:}\added{Profile shapes are consistent with the literature in every domain, with single-trait dominance for Grit and Leadership:} $C$ dominant for Grit, $E$
dominant for Leadership, $N$ most negative for Self-esteem alongside substantial $E$ and $C$
contributions matching the three-trait pattern in \citealt{robins2001}, and $C$-dominant with
secondary $N$ for Procrastination \citep{steel2007}. Values are means across ten checkpoints;
error bars show cross-checkpoint SD.}
\label{fig:criterion-betas}
\end{figure}

Table~\ref{tab:betas} reports the standardized $\beta$ coefficients across all trained models (Figure~\ref{fig:criterion-betas}). The discriminant evidence is strongest here, because -- unlike the zero-order profile, which a pure evaluative halo can already reproduce (Table~\ref{tab:criterion-baselines}) -- a halo cannot produce single-trait $\beta$ dominance. Across all model $\times$ domain cells the dominant-trait $\beta$ carries the correct sign in every case, and it is the largest-magnitude $\beta$ (argmax) in \deleted{27 of 36}\added{31 of 40} model$\times$domain cells (\added{77.5\%}); but this figure splits cleanly by domain into two strong single-trait cases and two weaker multi-trait cases, as the three patterns below make explicit.
For Grit and Leadership the $\beta$ picture is unambiguous: the dominant trait is the $\beta$-argmax in all checkpoints (\deleted{9/9}\added{10/10} each; C for Grit, E for Leadership), with the target coefficient at least 0.10 larger in magnitude than the next-largest. These are single-trait dominance patterns that a generic halo cannot generate, and they match the respective \deleted{meta-analytic}\added{published} findings; it is these two domains that carry the discriminant-validity claim.
For Procrastination the evidence is weaker: $\beta_C$ dominates $\beta_N$ in the mean ($-$0.37 vs +0.35), matching \citet{steel2007}'s finding that Conscientiousness is the strongest predictor and Neuroticism secondary, but Conscientiousness is the per-checkpoint $\beta$-argmax in only \deleted{5 of 9}\added{6 of 10} checkpoints, so we do not claim clean single-trait dominance here.
For Self-esteem the evidence is likewise weaker but interpretable: the $\beta$ pattern reproduces the three-trait loading reported by \citet{robins2001}, with N most negative alongside substantial and comparable E and C contributions. $\beta_N$ CIs exclude zero on every checkpoint (mean $\beta_N$ = $-$0.29 $\pm$ 0.08), so Neuroticism's unique contribution is statistically isolated throughout; but N is the $\beta$-argmax in only \deleted{4 of 9}\added{5 of 10} checkpoints, reflecting the near-equality of |$\beta_N$|, |$\beta_E$|, and |$\beta_C$|. We read this as the published multi-trait structure for self-esteem rather than clean discriminant dominance — consistent with the construct, but not, on its own, decisive against a halo account.

\begin{table}[H]
\centering
\caption{Standardized $\beta$ coefficients from multiple regression of each domain score on all five Big Five predictors simultaneously (z-scored inputs and outcomes). Values are mean $\pm$ SD across all \deleted{nine}\added{ten} trained checkpoints (five \deleted{Qwen seeds and four Llama-3-8B-4bit seeds}\added{seeds per backbone}). Target: the theoretically dominant trait and direction per \deleted{meta-analytic}\added{the published} literature. Argmax match: fraction of the \deleted{nine}\added{ten} checkpoints where the largest-magnitude $\beta$ corresponds to the target trait. R$^2$: mean variance explained. The target-trait $\beta$ CI excludes zero in every domain.}
\label{tab:betas}
\resizebox{\textwidth}{!}{%
\begin{tabular}{lcccccccc}
\toprule
Domain & Target & $\beta_O$ & $\beta_C$ & $\beta_E$ & $\beta_A$ & $\beta_N$ & Argmax match & R$^2$ \\
\midrule
Grit & +C & +0.16 $\pm$ 0.03 & +0.36 $\pm$ 0.06 & +0.14 $\pm$ 0.04 & +0.14 $\pm$ 0.03 & $-$0.18 $\pm$ 0.05 & \deleted{9/9}\added{10/10} & 0.43 \\
Self-esteem & $-$N & +0.17 $\pm$ 0.04 & +0.27 $\pm$ 0.05 & +0.29 $\pm$ 0.02 & +0.16 $\pm$ 0.04 & $-$0.29 $\pm$ 0.08 & \deleted{4/9}\added{5/10} & 0.57 \\
Leadership & +E & +0.25 $\pm$ 0.02 & +0.26 $\pm$ 0.05 & +0.44 $\pm$ 0.06 & +0.16 $\pm$ 0.04 & +0.08 $\pm$ 0.05 & \deleted{9/9}\added{10/10} & 0.57 \\
Procrastination & $-$C & $-$0.06 $\pm$ 0.04 & $-$0.37 $\pm$ 0.06 & $-$0.22 $\pm$ 0.05 & $-$0.07 $\pm$ 0.02 & +0.35 $\pm$ 0.06 & \deleted{5/9}\added{6/10} & 0.53 \\
\bottomrule
\end{tabular}
}
\end{table}

\added{\textbf{Direct human validation on identical instruments (Grit).} The analyses above compare silicon correlates to \deleted{meta-analytic}\added{published} \textit{targets}. For one domain we can make the comparison direct, within a single human dataset. The openpsychometrics.org Grit Scale dataset\footnote{\url{https://openpsychometrics.org/_rawdata/duckworth-grit-scale-data.zip}} administered the full IPIP-50 Big Five pool --- the same item pool the simulator is built on --- together with the 12-item Grit Scale to the same respondents ($n = 4{,}029$ complete cases), and the eight Grit-S items \citep{duckworth2009} map one-to-one onto the Grit-Scale items. Human and silicon can therefore be scored on exactly the same 20 Big Five marker items and eight Grit-S items, with the identical reverse-keying and standardized-$\beta$ pipeline (Table~\ref{tab:grit-human}). The dominant Conscientiousness relationship is recovered \deleted{almost exactly}\added{closely} --- human $r_C = {+}0.56$ vs.\ silicon ${+}0.54$; human $\beta_C = {+}0.50$ vs.\ silicon ${+}0.39$ --- with Conscientiousness the largest coefficient in both the zero-order and the partial ($\beta$) solution for humans and silicon alike, and Neuroticism matched in sign and magnitude ($-0.32$ vs.\ $-0.33$). The five-element profiles are highly concordant: the human-vs-silicon profile correlation is $0.93$ for the zero-order $r$-vector and $0.94$ for the $\beta$-vector. The one systematic discrepancy is the residual evaluative halo discussed above --- silicon inflates the off-target desirable traits relative to humans, most visibly Openness ($r = {+}0.37$ silicon vs.\ ${+}0.08$ human) --- so the silicon zero-order vector is shifted uniformly upward while preserving the correct rank order and dominant trait. This is a direct human benchmark for a silicon criterion-validity claim, rather than a comparison to tabulated \deleted{meta-analytic}\added{published} values, and it corroborates the construct$\rightarrow$response reading on the one domain for which a matched human dataset is available.}

\begin{table}[H]
\centering
\caption{\added{Direct human vs.\ silicon comparison for Grit on identical instruments. Human: openpsychometrics.org Grit Scale dataset (IPIP-50 Big Five $+$ 12-item Grit Scale), $n = 4{,}029$ complete cases. Silicon: Qwen criterion battery, five seeds, $n = 1000$, $T = 1.0$. Both are scored on the same 20 Big Five marker items and eight Grit-S items with identical reverse-keying. $r$: zero-order Pearson correlation of the Grit score with each observed Big Five scale; $\beta$: standardized coefficient from the simultaneous regression Grit $\sim O+C+E+A+N$. Human brackets are 95\% bootstrap CIs; silicon is mean $\pm$ SD across seeds. Literature $r$ from \citet{duckworth2009}. Human regression $R^2 = 0.41$.}}
\label{tab:grit-human}
\resizebox{\textwidth}{!}{%
\begin{tabular}{lccccc}
\toprule
Trait & Literature $r$ & Human $r$ [95\% CI] & Silicon $r$ (mean$\pm$SD) & Human $\beta$ [95\% CI] & Silicon $\beta$ (mean$\pm$SD) \\
\midrule
O & $+0.06$ & $+0.08$ $[+0.05, +0.11]$ & $+0.37 \pm 0.05$ & $+0.02$ $[-0.00, +0.05]$ & $+0.16 \pm 0.03$ \\
C & $+0.77$ & $+0.56$ $[+0.53, +0.58]$ & $+0.54 \pm 0.02$ & $+0.50$ $[+0.48, +0.53]$ & $+0.39 \pm 0.04$ \\
E & $+0.20$ & $+0.22$ $[+0.19, +0.25]$ & $+0.30 \pm 0.06$ & $+0.08$ $[+0.05, +0.11]$ & $+0.13 \pm 0.05$ \\
A & $+0.24$ & $+0.30$ $[+0.27, +0.33]$ & $+0.35 \pm 0.03$ & $+0.17$ $[+0.14, +0.20]$ & $+0.12 \pm 0.02$ \\
N & $-0.40$ & $-0.32$ $[-0.35, -0.30]$ & $-0.33 \pm 0.07$ & $-0.19$ $[-0.22, -0.17]$ & $-0.15 \pm 0.03$ \\
\bottomrule
\end{tabular}}
\end{table}

\subsection{Effects of Temperature on the Silicon Samples}
\label{sec:temperature}

\deleted{Here we analyze how the temperature, the fundamental parameter to control the level of stochasticity of the LLMs, influence the sampling generations.}\added{This section examines how the sampling temperature $T$ (Equation~\ref{eq:readout}), which controls how peaked each response distribution is, affects the silicon samples.}

For each of \added{the ten} trained models $\times$ two inventories (IPIP, BFI), the human respondent pool is split 50/50 by deterministic permutation into a calibration half and a reporting half. Silicon responses are generated from scratch at nine temperatures ($T \in \{0.4, 0.5, \ldots, 1.2\}$) against the full silicon population (n = 1000). The calibration half selects $T^*$ by minimising a moment-match objective: the per-item root mean squared difference between the human and silicon response distributions in their mean and standard deviation. The reporting half is used for all the metrics and never informs T selection.

Table~\ref{tab:temp-calib} shows the calibration-optimal $T^*$ and the calibration objective value at $T^*$, aggregated (mean $\pm$ SD) across seeds. In all four backbone $\times$ inventory configurations, mean $T^*$ falls at 1.12–1.20  with SD $\leq$ 0.05, i.e. essentially every individual seed selects the same calibration-optimal temperature. \added{Because $T^* = 1.20$ coincides with the upper edge of the grid in three of four configurations, and the objective is still decreasing there, the true moment-matching optimum may lie above 1.2.} The calibration objective decreases monotonically across the grid: at low temperatures silicon per-item response distributions are over-peaked relative to humans (the model concentrates probability more sharply on a few Likert options than humans do), and raising T broadens the marginal until the silicon spread matches human spread.

\begin{table}[ht!]
\centering
\caption{Calibration-optimal temperature $T^*$ selected by moment-match on the calibration half of human respondents, aggregated across seeds (mean $\pm$ SD). The calibration objective is per-item mean+std RMSE between human and silicon response distributions (lower is better).}
\label{tab:temp-calib}
\resizebox{\textwidth}{!}{%
\begin{tabular}{lccccc}
\toprule
Backbone & Inventory  & Mean $T^*$ & SD $T^*$ & Mean calib. obj at $T^*$ & SD calib. obj \\
\midrule
Qwen & IPIP & 1.20 & 0.00 & 0.537 & 0.034 \\
Qwen & BFI  & 1.12 & 0.05 & 0.394 & 0.022 \\
Llama & IPIP & 1.20 & 0.00 & 0.518 & 0.054 \\
Llama & BFI & 1.20 & 0.00 & 0.421 & 0.043 \\
\bottomrule
\end{tabular}
}
\end{table}

Table~\ref{tab:temp-tucker} shows Tucker congruence on the reporting half across the full temperature grid, aggregated across seeds. The key observation is that Tucker varies little with T: the aggregate mean spread across T $\in$ [0.4, 1.2] ranges from 0.015 $\pm$ 0.007 (Qwen IPIP) to 0.046 $\pm$ 0.008 (Llama IPIP). Factor structure is essentially T-invariant over this range. Intuitively, Tucker congruence depends on item-item correlations, a correlation is unchanged by an affine rescaling of either variable, and within this range T acts approximately as such a rescaling of each item's marginal spread, leaving the across-respondent co-variation that the structure depends on largely intact. The near-invariance is empirical and local to this range rather than exact. \added{Values at $T = 1.0$ differ slightly from Tables~\ref{tab:ipip-heldout}--\ref{tab:bfi-ood} (e.g., Qwen $\times$ IPIP 0.872 vs.\ 0.850) because the human reference here is the reporting half only, and the silicon responses were regenerated with fresh sampling draws.}

\begin{table}[ht!]
\centering
\caption{Tucker congruence on the reporting half of human respondents across five representative temperatures, mean $\pm$ SD across seeds. Mean spread = mean across seeds of (max $-$ min) across the full grid T $\in$ [0.4, 1.2].}
\label{tab:temp-tucker}
\resizebox{\textwidth}{!}{%
\begin{tabular}{lcccccc}
\toprule
Backbone $\times$ inventory & T=0.4 & T=0.6 & T=0.8 & T=1.0 & T=1.2 & Mean spread \\
\midrule
Qwen $\times$ IPIP & 0.861 $\pm$ 0.010 & 0.867 $\pm$ 0.006 & 0.870 $\pm$ 0.009 & 0.872 $\pm$ 0.003 & 0.868 $\pm$ 0.010 & 0.015 $\pm$ 0.007 \\
Qwen $\times$ BFI & 0.819 $\pm$ 0.010 & 0.815 $\pm$ 0.009 & 0.812 $\pm$ 0.013 & 0.808 $\pm$ 0.017 & 0.809 $\pm$ 0.014 & 0.021 $\pm$ 0.007 \\
Llama $\times$ IPIP & 0.900 $\pm$ 0.035 & 0.898 $\pm$ 0.029 & 0.888 $\pm$ 0.029 & 0.877 $\pm$ 0.033 & 0.861 $\pm$ 0.035 & 0.046 $\pm$ 0.008 \\
Llama $\times$ BFI & 0.815 $\pm$ 0.043 & 0.817 $\pm$ 0.044 & 0.807 $\pm$ 0.052 & 0.801 $\pm$ 0.048 & 0.800 $\pm$ 0.056 & 0.035 $\pm$ 0.013 \\
\bottomrule
\end{tabular}
}
\end{table}

At the calibration-optimal $T^*$, the aggregate mean Tucker change relative to the \deleted{prior eval-tuned T}\added{temperature used for Tables~\ref{tab:ipip-heldout}--\ref{tab:bfi-ood} ($T = 1.0$ for both backbones)} is small in every cell. Computed as Tucker at $T^*$ minus Tucker at $T = 1.0$ from Table~\ref{tab:temp-tucker}: Qwen IPIP $\Delta$ = $-$0.004, Qwen BFI $\Delta$ $\approx$ 0.000, Llama IPIP $\Delta$ = $-$0.016, and Llama BFI $\Delta$ = $-$0.001, so every $|\Delta| \leq 0.016$ --- small relative to the cross-seed SDs of Tables~\ref{tab:ipip-heldout}--\ref{tab:bfi-ood}.

Temperature therefore plays two distinct roles that do not interact. It is a marginal-distribution calibration knob: changing T adjusts per-item response spread so that silicon distributions resemble human distributions, and the calibration objective selects the T that minimises this mismatch. Separately, it is not a factor-structure knob: the Tucker congruence \deleted{and criterion validity that constitute our central claims are}\added{that constitutes our central structural claim is} robust across the full temperature grid and \deleted{do}\added{does} not depend on the specific T chosen during initial evaluation. \added{Criterion validity was evaluated only at $T = 1.0$; its sensitivity to temperature remains untested.}

\subsection{Generative stability and internal consistency}
\label{sec:stability}

We report two complementary measures: generative (sampling) stability across repeated administrations and internal consistency (Cronbach $\alpha$) within a single administration. We deliberately avoid the term test-retest reliability here: human test-retest additionally reflects occasion and trait/state variation over time, whereas our procedure fixes the latent and re-runs only the categorical draw, isolating generative noise alone.

The respondent population is held fixed at n = 500 latents z drawn once and saved. Each instrument is administered twice\added{ at $T = 1.0$} under different response-sampling seeds, keeping z, items, and checkpoint identical. The two administrations therefore differ only in the categorical sampling noise applied to each item's digit distribution — the sole stochastic step in silicon\_sample. Correlating reverse-coded scale scores across the 500 respondents gives a per-factor Pearson r that measures generative stability: how reproducibly the same latent z maps to the same scale score when the sampling is rerun.

All trained models are evaluated (five \deleted{Qwen seeds and four Llama seeds; the Qwen aggregates exclude the bad-init seed 3, as elsewhere}\added{seeds per backbone}), each on IPIP and a BFI short form. Tables~\ref{tab:reliability-r} and~\ref{tab:reliability-alpha} report per-factor generative-stability r and Cronbach $\alpha$ respectively, aggregated as mean $\pm$ SD across seeds within each model. Administrations 1 and 2 produced \deleted{near-identical $\alpha$ values in all cells (within 0.06)}\added{$\alpha$ values differing by at most 0.06 in any cell}, confirming internal consistency is stable across sampling noise.

\begin{table}[ht!]
\centering
\caption{Silicon generative stability: mean $\pm$ SD across seeds of the Pearson r between reverse-coded scale scores from two independent administrations with the same 500 fixed latents z per configuration. Values reflect generative-stability reliability — variability from categorical sampling noise only. O = Openness, C = Conscientiousness, E = Extraversion, A = Agreeableness, N = Neuroticism. Mean = unweighted mean across the five factors. The IPIP scales use the four held-out items per factor; the BFI scales here use a 25-item BFI short form (five items per factor) administered for this reliability analysis only, distinct from the 60-item BFI-2 of \S\ref{sec:bfi}.}
\label{tab:reliability-r}
\resizebox{\textwidth}{!}{%
\begin{tabular}{lcccccc}
\toprule
Config & O & C & E & A & N & Mean \\
\midrule
Qwen $\times$ IPIP & 0.52 $\pm$ 0.05 & 0.61 $\pm$ 0.03 & 0.80 $\pm$ 0.02 & 0.55 $\pm$ 0.06 & 0.66 $\pm$ 0.04 & 0.63 \\
Qwen $\times$ BFI & 0.57 $\pm$ 0.07 & 0.59 $\pm$ 0.02 & 0.77 $\pm$ 0.01 & 0.71 $\pm$ 0.03 & 0.82 $\pm$ 0.00 & 0.69 \\
Llama $\times$ IPIP & 0.51 $\pm$ 0.04 & 0.51 $\pm$ 0.03 & 0.72 $\pm$ 0.03 & 0.69 $\pm$ 0.02 & 0.71 $\pm$ 0.04 & 0.63 \\
Llama $\times$ BFI & 0.61 $\pm$ 0.02 & 0.55 $\pm$ 0.04 & 0.70 $\pm$ 0.02 & 0.73 $\pm$ 0.04 & 0.85 $\pm$ 0.01 & 0.69 \\
\bottomrule
\end{tabular}
}
\end{table}

\begin{table}[ht!]
\centering
\caption{Cronbach $\alpha$: mean $\pm$ SD across seeds. Scale lengths: IPIP the 4 held-out items per factor; BFI a 25-item short form, 5 items per factor (the reliability-only set of Table~\ref{tab:reliability-r}, distinct from the 60-item BFI-2 of \S\ref{sec:bfi}).\added{ Human rows: $\alpha$ for the same items computed in the human samples.}}
\label{tab:reliability-alpha}
\resizebox{\textwidth}{!}{%
\begin{tabular}{lcccccc}
\toprule
Config & O & C & E & A & N & Mean \\
\midrule
Qwen $\times$ IPIP & 0.35 $\pm$ 0.06 & 0.54 $\pm$ 0.02 & 0.76 $\pm$ 0.02 & 0.47 $\pm$ 0.06 & 0.57 $\pm$ 0.06 & 0.54 \\
Qwen $\times$ BFI & 0.46 $\pm$ 0.08 & 0.54 $\pm$ 0.02 & 0.72 $\pm$ 0.01 & 0.73 $\pm$ 0.03 & 0.82 $\pm$ 0.00 & 0.65 \\
Llama $\times$ IPIP & 0.39 $\pm$ 0.06 & 0.45 $\pm$ 0.05 & 0.64 $\pm$ 0.05 & 0.63 $\pm$ 0.02 & 0.65 $\pm$ 0.03 & 0.55 \\
Llama $\times$ BFI & 0.51 $\pm$ 0.04 & 0.51 $\pm$ 0.04 & 0.66 $\pm$ 0.03 & 0.73 $\pm$ 0.04 & 0.84 $\pm$ 0.01 & 0.65 \\
\added{\midrule}
\added{Human $\times$ IPIP} & \added{0.52} & \added{0.70} & \added{0.81} & \added{0.66} & \added{0.75} & \added{0.69} \\
\added{Human $\times$ BFI} & \added{0.61} & \added{0.74} & \added{0.77} & \added{0.72} & \added{0.82} & \added{0.73} \\
\bottomrule
\end{tabular}
}
\end{table}

Four findings emerge. First, silicon generative-stability r and Cronbach $\alpha$ are of broadly similar magnitude to values reported for published short-form Big Five inventories in human samples (test-retest r typically 0.65–0.85; $\alpha$ typically 0.50–0.85 for 4–5 item scales); the mean stability r across all four configurations ranges from 0.63 to 0.69. We stress that this is an orienting comparison, not a construct-equivalent one: human test-retest reliability additionally reflects occasion and trait/state variation across sessions, whereas our quantity fixes the latent and varies only the categorical sampling, so it isolates generative noise \deleted{and is itself tunable via temperature (Finding 4; {\S\ref{sec:stability}})}\added{and depends on the sampling temperature, held fixed at $T = 1.0$ here}. The similarity therefore indicates that the population is neither deterministic nor unstable — not that the simulator reproduces human test-retest reliability per se. Second, the cross-seed SDs are small for most factors, indicating that generative stability is a consistent property of the training process. The largest variance appears on Qwen BFI Openness (0.57 $\pm$ 0.07) and Qwen IPIP Agreeableness (0.55 $\pm$ 0.06)\deleted{, factors that also show wider structural variance in {\S\ref{sec:heldout}}–{\S\ref{sec:bfi}}}. \deleted{Third, the factor-level ordering is consistent across both backbones and both measures: Extraversion and Neuroticism are the most stable factors, Openness is the least stable on IPIP, and Conscientiousness joins Openness at the bottom on BFI. This mirrors the reliability ordering seen in human short-form inventories.}\added{Third, the factor-level ordering is broadly similar across backbones and measures: Neuroticism is among the two most stable factors in every configuration, Extraversion in most (Agreeableness overtakes it in three of the four BFI cells), and Openness is the least or tied-least stable factor on IPIP, joined by Conscientiousness at the bottom on BFI. Whether this matches the human ordering can be checked against the human $\alpha$ on the same items (human reference rows at the bottom of Table~\ref{tab:reliability-alpha}). The human $\alpha$ ordering largely matches: Openness is the least internally consistent factor in both human inventories and Neuroticism (with Extraversion) among the most, mirroring the silicon pattern; the main exception is Agreeableness, which is among the least consistent human scales but among the most stable in silicon on the BFI.} Fourth, Qwen and Llama show nearly identical mean stability r (both 0.63 on IPIP, both 0.69 on BFI), \deleted{indicating that generative stability is determined primarily by the instrument and temperature rather than by LLM architecture.}\added{suggesting that, at fixed temperature, generative stability depends more on the instrument than on the backbone.}

\section{Discussion}

The results \deleted{described in section {\S{}5}}\added{of Section~\ref{sec:results}} form a chain of evidence rather than a single number, and the chain is what the argument rests on. Taken individually, any one link could be explained away: a Tucker congruence of 0.85 on held-out items could reflect memorized item-response patterns rather than a learned construct; transfer to the BFI could reflect surface overlap between the two inventories' wording; the criterion correlations could reflect a generic evaluative halo; the multi-seed numbers could reflect a single lucky hyperparameter setting; the temperature used to generate every other table could itself be a tuned artifact. Taken together, the links close off these explanations one at a time. Cross-inventory transfer rules out item memorization, since the BFI's 60 items share no text with the IPIP training items; a model that had merely memorized IPIP item-response statistics would have no mechanism for producing the right factor structure on wording it has never seen. The criterion battery rules out replay of the injected covariance $\Sigma_{\text{human}}$, since the domain–trait relationships tested there — grit with conscientiousness, self-esteem with neuroticism, leadership with extraversion, procrastination with conscientiousness — are absent from $\Sigma_{\text{human}}$ by construction; the training procedure injects only the Big Five inter-factor structure, never a personality–outcome link, so a model that recovers \deleted{those links with the correct sign in all checkpoint–domain cells, and the published multi-trait $\beta$ pattern in the cases studied closely (self-esteem, procrastination),}\added{the dominant-trait link with the correct sign in every checkpoint--domain cell} is doing something beyond covariance replay. \deleted{The discriminant-validity analysis in particular shows that the apparent zero-order halo for self-esteem and procrastination was a property of the zero-order correlations, not of the underlying representation: once the five traits are regressed on simultaneously, the unique contribution of the target trait is statistically isolated in every case, and the resulting profile matches the multi-trait pattern reported in the literature rather than a generic positivity loading.}\added{The discriminant analysis sharpens this: with all five traits entered simultaneously, Grit and Leadership show single-trait dominance in every checkpoint, which a generic evaluative halo cannot produce. For self-esteem and procrastination the target trait's unique contribution is statistically isolated but does not consistently dominate, and the direct human comparison for Grit (Table~\ref{tab:grit-human}) shows that a residual halo remains on off-target desirable traits; these two domains are consistent with, but not decisive for, the construct$\rightarrow$response account.} \deleted{The nine-checkpoint replication rules out single-run or single-backbone luck, since the headline numbers reproduce within a few hundredths of Tucker congruence across all Qwen seeds and Llama-3-8B seeds — a substantially larger model with a different tokenizer, architecture, and pretraining corpus.}\added{The multi-seed, two-backbone replication makes single-run or single-backbone luck unlikely: mean congruence differs by at most 0.023 between backbones on either inventory, although cross-seed variability is appreciably larger for Llama-3-8B (SD up to 0.074 on the BFI-2).} The held-out temperature calibration rules out the headline numbers being an artifact of hand-tuning T against the same data used to report Tucker congruence, since selecting T on a disjoint calibration half of the human sample and reporting on the other half changes Tucker congruence by at most a few hundredths in either direction, \deleted{and the calibration-optimal temperature is consistently higher than the temperature originally chosen by inspection — meaning the original choice, if anything, understated rather than inflated the headline numbers.}\added{and in three of four configurations the calibration-optimal temperature yields slightly lower congruence (largest change: Llama IPIP, $-0.016$), so the inspection-chosen temperature, if anything, slightly flattered the headline numbers, by an amount small relative to cross-seed variability.}

The reliability analysis addresses a different kind of concern: even a genuinely construct-valid synthetic population is not useful for downstream survey methodology if its measurement properties look nothing like a real instrument's. Silicon generative-stability correlations and Cronbach alphas are of broadly similar magnitude to those reported for published human short-form Big Five inventories (an orienting rather than construct-equivalent comparison, since our quantity isolates sampling noise and \deleted{is temperature-tunable), and the factor-level ordering of stability — extraversion and neuroticism most stable, openness least — mirrors the ordering of reliability seen in human samples.}\added{depends on the sampling temperature), and the factor-level ordering of stability is broadly consistent across backbones and instruments (Section~\ref{sec:stability}).} A simulator that reproduced factor structure but had, say, near-perfect or near-zero generative stability would be a poor stand-in for a human population even if its Tucker congruence looked identical on paper; the fact that the noise properties also resemble a real instrument is part of the validity case, not a separate result.

\deleted{It is worth to note that the gate ablation  shows that the latent gating mechanism we built into the architecture — a sigmoid gate over the LoRA bottleneck, modulated by the trait code — does not produce so much difference: gated and ungated variants differ by about 0.02 Tucker congruence in opposite directions on IPIP versus BFI.}\added{The gate ablation shows that the latent gating mechanism --- a sigmoid gate over the LoRA bottleneck, modulated by the trait code --- makes no practical difference: gated and ungated variants differ by 0.014--0.016 Tucker congruence, in opposite directions on IPIP and BFI-2.} We report this transparently because it reallocates where we believe the paper's contribution actually lives: not in the gating mechanism, but in the combination of continuous soft-prefix and LoRA conditioning, the sampling-based (rather than argmax) construction of the silicon population, and the validity battery itself.

Positioned against the broader silicon-sampling literature, this work shares the goal of \citet{argyle2023}'s algorithmic-fidelity argument and the persona-prompting line of work \citep{serapiogarcia2023,jiang2023} — using language models to stand in for human respondents — but targets a different and, we think, more demanding bar: not whether prompted or conditioned trait scores correlate with intended traits, but whether a synthetic population built from such a model reproduces the factor structure, cross-instrument transfer, and nomological network that define a construct in the psychometric sense, and does so with the kind of statistical replication and robustness checks the methodology requires before the resulting synthetic data could be used for actual scale development or pilot testing. Relative to the persona-prompting baselines, trained latent conditioning nearly doubles factor congruence over the best zero-shot prompt format on the held-out items (with a smaller but still substantial margin out of distribution), which suggests that whatever silicon samples are capturing, it depends on more than the backbone's pre-existing capacity to role-play a description in context.

These results also suggest a range of practical applications, provided silicon samples are used to complement human data rather than replace it. Because the simulator defines a response distribution over the full latent continuum, candidate items for new or revised scales could be administered to a silicon population before fielding. Items with weak or cross-loading patterns, low item–total correlations, or a negative effect on internal consistency could then be flagged and revised at negligible cost, and the cross-inventory transfer of \deleted{§5.2}\added{Section~\ref{sec:bfi}} suggests this screening is not limited to items resembling the training inventory. The same population, which reproduces \added{the sign pattern of} human inter-factor correlations and \added{broadly} realistic levels of reliability, offers a more plausible data-generating process than idealized covariance matrices for sample-size planning and for dry runs of preregistered analysis pipelines. Because z is an explicit input, researchers can also oversample rare trait combinations or the tails of the distribution to probe ceiling and floor effects that are costly to study with recruited participants. Finally, administering new instruments to a silicon population yields predicted Big Five profiles that can help prioritize which personality–outcome relationships merit human testing, and populations generated from summary statistics alone could support teaching and software benchmarking. These uses have clear limits: silicon samples should not serve for norming, for decisions about real individuals, or as the sole evidence in confirmatory work, and instruments with substantial negative-affect content warrant extra caution given the weaker reproduction of Neuroticism. We therefore see their natural role as a pre-screening step, followed by a human pilot and a full human study for any substantive conclusion.

\section{Conclusion}

This paper describes and validates a method for building synthetic survey respondents that reproduce human Big Five psychometric structure: a frozen instruction language model conditioned on a continuous five-dimensional trait latent through a soft prefix and \added{(optionally)} gated low-rank adapters, with each Likert response read directly from the model's own digit-token logits and sampled, rather than taken as an argmax, so that a synthetic population can be built whose item-level variability resembles a real respondent pool. Across \deleted{nine}\added{ten} trained checkpoints spanning two backbones (five \deleted{Qwen seeds and four Llama-3-8B seeds}\added{seeds each}), the resulting silicon populations reproduce the human five-factor structure on items withheld from training, transfer that structure to a different inventory with different item wording and format, \deleted{preserve the inter-factor correlation matrix}\added{preserve the sign pattern of the inter-factor correlations}, and — most importantly for the validity argument — recover documented relationships between personality and four non-personality constructs that were never injected into the training procedure, \deleted{with the correct sign in every one of the checkpoint–domain cells and literature-matching multi-trait loadings under discriminant analysis.}\added{with the dominant trait carrying the correct sign in every checkpoint--domain cell, single-trait dominance for grit and leadership under discriminant analysis, and multi-trait patterns consistent with the literature for self-esteem and procrastination.}

These results make a reasonably strong case that the simulator is doing more than memorizing item statistics or replaying an injected correlation matrix, but several limitations bound how far that case extends. The conditioning latents used throughout are CFA factor scores derived from the same IPIP sample the model is trained on, and the inter-factor covariance $\Sigma_{\text{human}}$ used to draw the synthetic population is injected directly, so a degree of circularity is built into the setup by design; the emergent criterion validity in \S\ref{sec:criterion} addresses this only partly, since it shows the recovered domain–trait links are not themselves injected, while the trait space they are expressed in still originates from the training inventory. \deleted{Relatedly, criterion validity throughout this paper is benchmarked against meta-analytic literature targets rather than against a human sample in which the Big Five and the criterion instruments were measured together; a direct human-versus-silicon comparison on co-administered instruments would be a stronger test than comparison against a published correlation, and is the most valuable single follow-up this work points toward.}\added{Relatedly, criterion validity is benchmarked against published literature targets for three of the four domains; only for grit could we compare silicon and human respondents on co-administered instruments (Table~\ref{tab:grit-human}), where the dominant Conscientiousness link was recovered closely but a residual evaluative halo remained. Extending matched human comparisons to the other domains is the most valuable single follow-up this work points toward.} Two factors are handled less well than the others, and for different reasons. Neuroticism has the lowest structural congruence on held-out items (0.74--0.76, below the ``fair'' threshold), and this is a genuine shortfall of the simulator rather than a limit of the construct or the analysis: on the identical items, two human subsamples recover Neuroticism at 0.98 (\deleted{{\S\ref{sec:heldout}}}\added{Section~\ref{sec:indices}}), so the model is failing to reproduce structure that is clearly recoverable in humans. Openness, by contrast, is weakest not on congruence --- where it is in fact among the stronger factors (0.83--0.90) --- but on internal consistency and generative stability \deleted{(lowest Cronbach $\alpha$ and lowest retest $r$ of the five factors)}\added{(lowest or tied-lowest Cronbach $\alpha$ in all four backbone $\times$ inventory configurations, and lowest or tied-lowest generative-stability $r$ in three of four)}; \deleted{because that weakness mirrors Openness's comparatively low internal consistency in human normative data, we read it as a property of the construct the method is faithfully tracking rather than a defect of the architecture.}\added{because human Openness scales on the same items also show comparatively low internal consistency (the lowest Cronbach $\alpha$ of the five factors in both human inventories; Table~\ref{tab:reliability-alpha}), we read this weakness as at least partly a property of the construct the method is tracking rather than solely a defect of the architecture.} We have not ruled out that targeted item augmentation could narrow either gap, and the Neuroticism gap in particular warrants it. Finally, the evidence base covers two inventories (IPIP and BFI-2) and a single construct family (the Big Five); whether the same conditioning approach and validity pattern would hold for other personality models, other psychological constructs, or instruments with substantially different item formats remains untested, and is the natural next axis along which to extend this work.

\section{Reproducibility}

All code, data splits, trained checkpoints, and per-checkpoint result tables are available at \url{https://github.com/milnico/SiliconRespondents}.

\section*{Open Practices Statement}

All data analyzed in this study are publicly available. The IPIP-50 Big Five responses are from the Open-Source Psychometrics Project (\url{https://openpsychometrics.org/_rawdata/}, archive \texttt{BIG5.zip}); the BFI-2 standardization sample is distributed with the \textit{ShinyItemAnalysis} R package \citep{martinkova2018}; and the Grit Scale dataset used for the direct human comparison is from the Open-Source Psychometrics Project (\url{https://openpsychometrics.org/_rawdata/duckworth-grit-scale-data.zip}). All code, data-partition scripts, trained model checkpoints, and per-checkpoint result tables required to reproduce every reported analysis are openly available at \url{https://github.com/milnico/SiliconRespondents}. None of the studies reported in this manuscript was preregistered.

\bibliography{references}

\end{document}